\documentclass[runningheads]{llncs}
\usepackage{graphicx}
\usepackage{comment}
\usepackage{amsmath,amssymb} 
\usepackage{color}

\graphicspath{ {./images/} }
\usepackage{epsfig}
\usepackage{float}
\usepackage{algorithm}
\usepackage{algorithmic}
\usepackage{amssymb}
\usepackage{booktabs}
\usepackage{cite}
\usepackage{subcaption}
\begin{document}
\pagestyle{headings}
\mainmatter
\def\ECCVSubNumber{4120}  

\title{DADA: Differentiable Automatic Data Augmentation}

\titlerunning{Differentiable Automatic Data Augmentation}
%
%

\author{Yonggang Li$^*$\inst{1} \and
	Guosheng Hu$^*$\inst{2,3} \and
	Yongtao Wang$^\dagger$\inst{1} \and
	Timothy Hospedales\inst{4} \and \\
	Neil M. Robertson\inst{2,3} \and
	Yongxin Yang\inst{4}}

\authorrunning{Y. Li et al.}
%
\institute{
	Wangxuan Institute of Computer Technology, Peking University, Beijing, China
	\email{wyt@pku.edu.cn}
	\and Anyvision, Queens Road, Belfast, United Kingdom
	\and Queens University of Belfast, Belfast, United Kingdom
	\and School of Informatics, The University of Edinburgh, Edinburgh, United Kingdom
}

\maketitle

\begin{abstract}
Data augmentation (DA) techniques aim to increase data variability, and thus train deep networks with better generalisation. The pioneering AutoAugment automated the search for optimal DA policies with reinforcement learning. However, AutoAugment is extremely computationally expensive, limiting its wide applicability. Followup works such as Population Based Augmentation (PBA) and Fast AutoAugment improved efficiency, but their optimization speed remains a bottleneck. In this paper, we propose Differentiable Automatic Data Augmentation (DADA) which dramatically reduces the cost. DADA relaxes the discrete DA policy selection to a differentiable optimization problem via Gumbel-Softmax. In addition, we introduce an unbiased gradient estimator, RELAX, leading to an efficient and effective one-pass optimization strategy to learn an efficient and accurate DA policy. We conduct extensive experiments on CIFAR-10, CIFAR-100, SVHN, and ImageNet datasets. Furthermore, we demonstrate the value of Auto DA in pre-training for downstream detection problems. Results show our DADA is at least one order of magnitude faster than the state-of-the-art while achieving very comparable accuracy. The code is available at https://github.com/VDIGPKU/DADA.

\keywords{AutoML, Data Augmentation, Differentiable Optimization}
\end{abstract}

\section{Introduction}
\label{sec:introduction}
\renewcommand{\thefootnote}{}
\footnotetext{$*$ indicates equal contribution. \\ $\dagger$ indicates corresponding author.}
{Data augmentation (DA) techniques, such as  geometric transformations (e.g., random horizontal flip, rotation, crop), color space augmentations (e.g., color jittering), are widely applied in training deep neural networks. DA serves as a regularizer to alleviate over-fitting by increasing the amount and diversity of the training data. Moreover, it is particularly important in scenarios where big training data is not available, e.g. medical image analysis.}


Data augmentation is very useful for training neural networks, however, it is nontrivial to {automatically} select  the most effective DA policy (a set of augmentation operations) for  one particular task and dataset. The pioneering work, AutoAugment (AA) \cite{DBLP:conf/cvpr/CubukZMVL19}, models {the process of policy selection} as an optimization problem: the objective is to maximize the accuracy on the validation set, the parameters optimized are i) the probabilities of {applying} different augmentation functions and ii) the magnitudes of chosen functions. Reinforcement learning is used to optimize this problem. AutoAugment achieved excellent performance on image classification, however, the optimization is very expensive: $\sim${5000} GPU hours for one augmentation search. Despite the effectiveness of AutoAugment, the heavy computational cost limits its value to most users. 
To address the efficiency problem, Population Based  Augmentation (PBA) \cite{DBLP:conf/icml/HoLCSA19} and Fast AutoAugment (Fast AA) \cite{DBLP:conf/nips/LimKKKK19} are proposed. PBA introduces an efficient population based optimization, which was originally used for hyper-parameter optimization. Fast AA models the data augmentation search as a density matching problem and solves it through bayesian optimization. 
Though PBA and Fast AA greatly improve search speed, augmentation policy learning remains rather slow, e.g. PBA still needs $\sim$5 GPU hours for one search on the reduced CIFAR-10 dataset.

\begin{figure}[t]
	\centering
	\begin{subfigure}[h]{0.49\linewidth}
		\includegraphics[width=\linewidth]{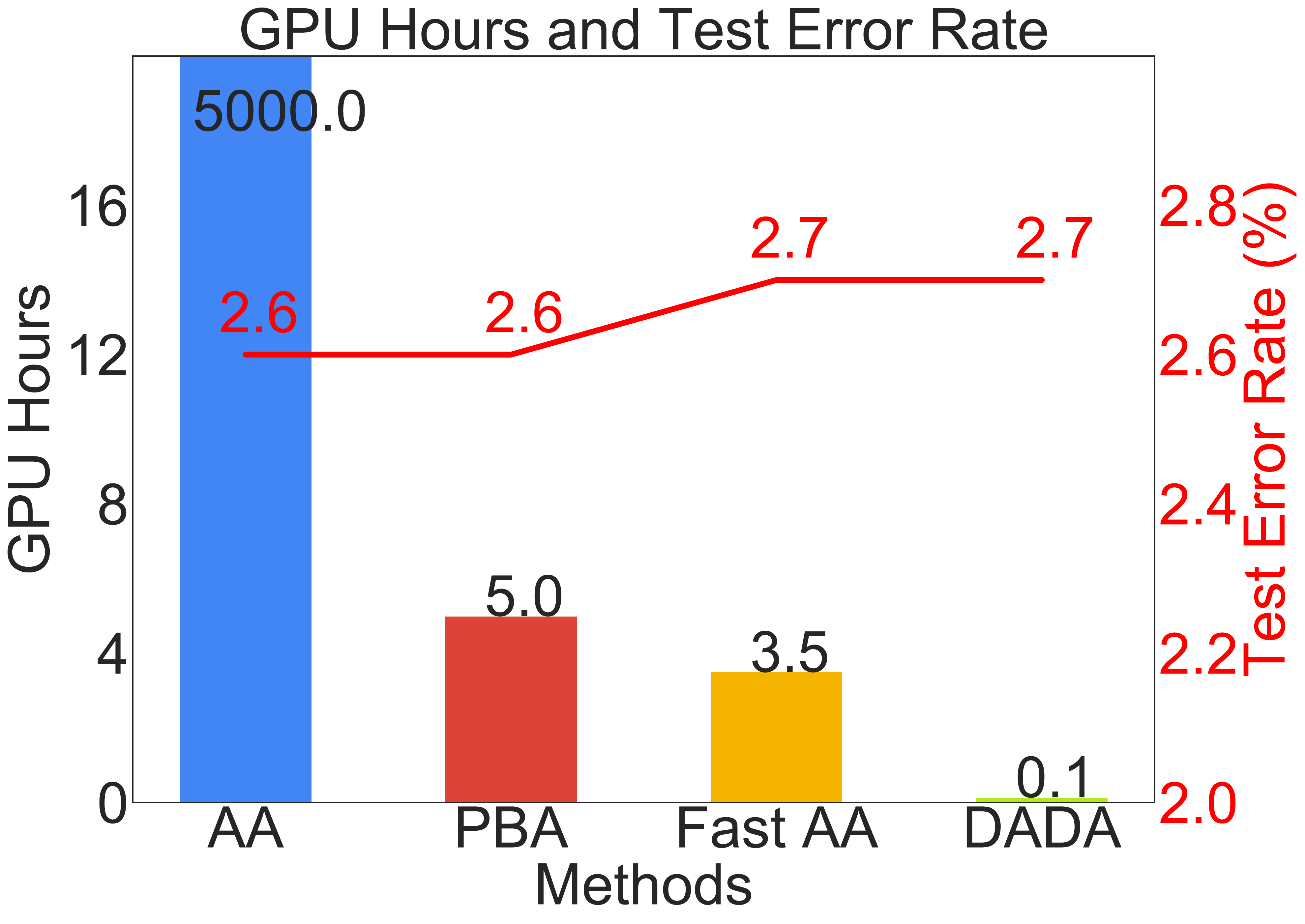}
		\caption{CIFAR-10}
		\label{subfig:timeperformances_cifar}
	\end{subfigure}
	\begin{subfigure}[h]{0.49\linewidth}
		\includegraphics[width=\linewidth]{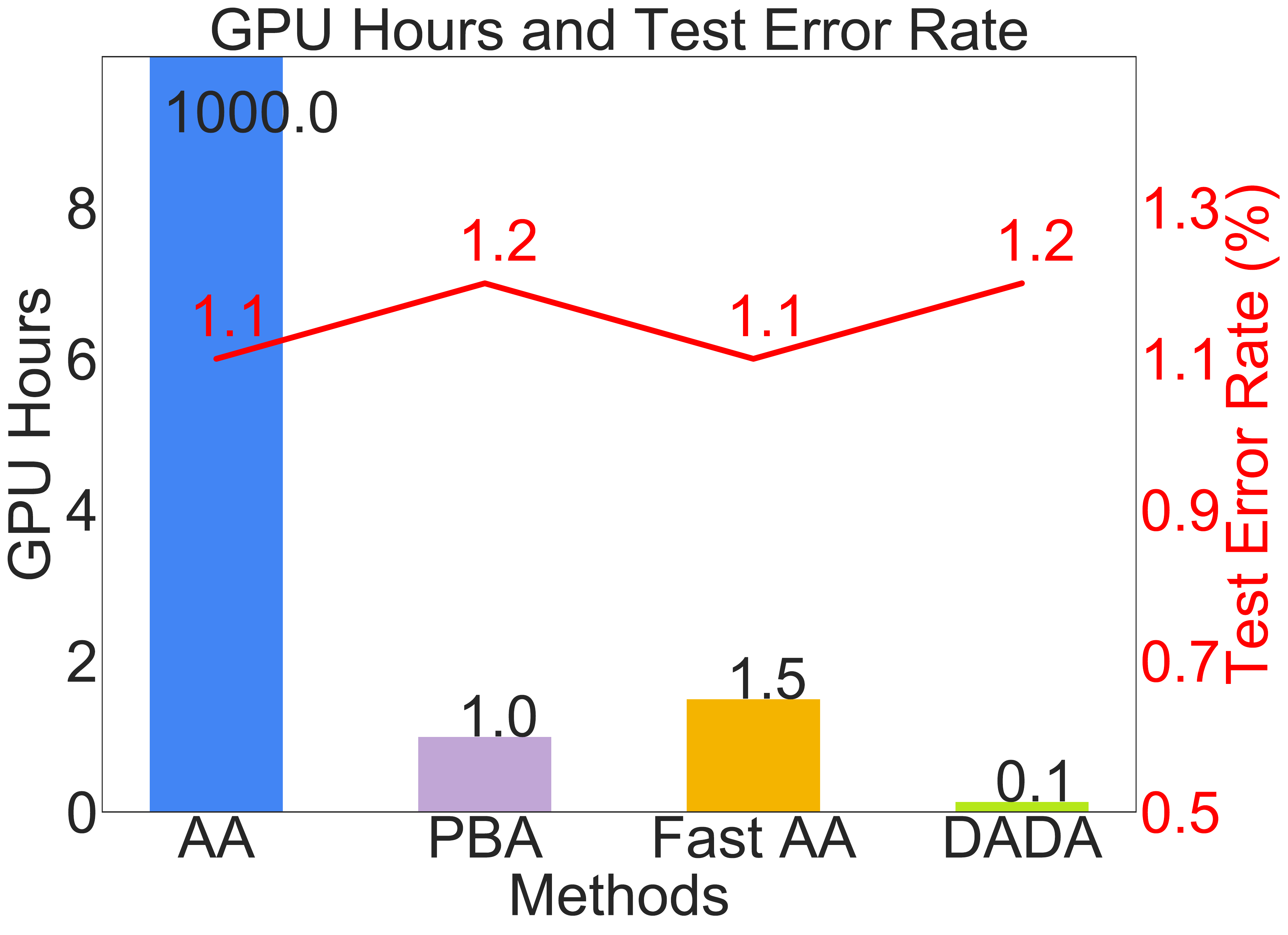}
		\caption{SVHN}
		\label{subfig:timeperformances_svhn}
	\end{subfigure}
	\caption{DADA achieves an order of magnitude reduction in computation cost while maintaining comparable test error rate to state of the art DA methods (AA \cite{DBLP:conf/cvpr/CubukZMVL19}, PBA \cite{DBLP:conf/icml/HoLCSA19} and Fast AA \cite{DBLP:conf/nips/LimKKKK19}) using WRN-28-10 backbone.
	}
	\label{fig:timeperformances}
\end{figure}

The inefficiency of the existing DA optimization strategies arises from the fact that the optimization {(selecting discrete augmentation functions)} is intrinsically non-differentiable, thus precluding joint optimization of network weights and DA parameters, and requiring resorting to multi-pass reinforcement learning, BayesOpt, and evolutionary strategies. 
Intuitively, optimization efficiency can be greatly improved if we can relax the optimization to a differentiable one and jointly optimize network weights and DA parameters in a single-pass way. Motivated by the differentiable neural architecture search \cite{DBLP:conf/iclr/LiuSY19,DBLP:conf/iclr/XieZLL19,Dong_2019_CVPR}, we propose a Differentiable Automatic Data Augmentation (DADA) to relax the optimization problem to be differentiable and then use gradient-based optimization to jointly train model weights and data augmentation parameters. In this way, we can achieve a very efficient and effective DA policy search.

DADA follows AA~\cite{DBLP:conf/cvpr/CubukZMVL19} in using a search space where a policy contains many sub-policies, each of which has two operations (each with probability and magnitude of applying the DA function).  DADA first reformulates the discrete search space to a joint distribution that encodes sub-policies and operations. Specifically, we treat the sub-policy selection and augmentation application as sampling from a Categorical distribution and Bernoulli distribution, respectively. In this way,  DA  optimization becomes a Monte Carlo gradient estimate problem~\cite{DBLP:journals/corr/abs-1906-10652}. However, Categorical and Bernoulli distributions are not differentiable. To achieve differentiable optimization, we relax the two non-differentiable distributions to be differentiable through Gumbel-Softmax gradient estimator~\cite{DBLP:conf/iclr/JangGP17} (a.k.a. concrete distribution~\cite{DBLP:conf/iclr/MaddisonMT17}). Furthermore, DADA minimizes the loss on validation set rather than the accuracy (used by AutoAugment) to facilitate gradient computation.

To realise this differentiable optimization framework, we introduce 1) an efficient optimization strategy and 2) an accurate gradient estimator. 
For 1), a straightforward solution for sampling-based optimization is to iterate two sub-optimizations until convergence: i) optimizing  DA policies ii) training neural networks. Clearly, this sequential  optimization is very slow. Motivated by DARTS~\cite{DBLP:conf/iclr/LiuSY19}, we \emph{jointly} optimize parameters of DA policies and networks through stochastic gradient descent. This \emph{one-pass} strategy greatly reduces the computational cost. For 2), the classic gradient estimator is the Gumbel-Softmax estimator. In the field of network architecture search (NAS), SNAS~\cite{DBLP:conf/iclr/XieZLL19} and GDAS~\cite{Dong_2019_CVPR} use this estimator and achieved good performance. However, the gradient estimated by Gumbel-Softmax estimator is biased. To overcome this, we propose to use the  RELAX~\cite{DBLP:conf/iclr/GrathwohlCWRD18} estimator, which can provide an unbiased gradient estimator unlike Gumbel-Softmax, and thus improved policy search.
We conduct extensive experiments using a variety of deep models and datasets, e.g.  CIFAR-10,  SVHN~\cite{netzer2011reading}. 
As shown in Fig.~\ref{fig:timeperformances}, our method achieves comparable performance with the state-of-the-art, while requiring significantly 
less compute.


The contributions of this work are threefold:
\begin{enumerate}
	\item We propose Differentiable Automatic Data Augmentation (DADA), which uses an efficient \emph{one-pass} gradient-based optimization strategy and achieves at least one order of magnitude speedup over state-of-the-art alternatives.
	\item DADA relaxes the  DA  parameter optimization  to be  differentiable via Gumbel-Softmax.  To achieve accurate gradient estimation,  
we introduce an unbiased gradient estimator, RELAX~\cite{DBLP:conf/iclr/GrathwohlCWRD18}, to  our DADA framework. 
	\item We perform a thorough evaluation on CIFAR-10, CIFAR-100~\cite{krizhevsky2009learning}, SVHN~\cite{netzer2011reading} and ImageNet~\cite{ILSVRC15}, as well as object detection benchmarks. We achieve at least an order of magnitude speedup over state-of-the-art while maintaining  accuracy. Specifically, on ImageNet, we only use 1.3 GPU (Titan XP) hours for searching and achieve 22.5\% top1-error rate with ResNet-50.
	
\end{enumerate}

\section{Related Work}
\label{sec:relatedwork}
In the past few years, handcrafted data augmentation techniques are widely used in training deep Deep Neural Network (DNN) models for image recognition, object detection, etc.
For example, rotation, translation, cropping, resizing, and flipping~\cite{lecun1998gradient, DBLP:journals/corr/SimonyanZ14a} are  commonly used to augment training examples.
Beyond these, there are other techniques manually designed with some domain knowledge, like Cutout\cite{DBLP:journals/corr/abs-1708-04552}, Mixup~\cite{DBLP:conf/iclr/ZhangCDL18}, and CutMix~\cite{Yun_2019_ICCV}.
Although these methods  achieve promising improvements on the corresponding tasks, they need expert knowledge to design the operations and set the hyper-parameters for specific datasets.

Recently, inspired by the neural architecture search (NAS) algorithms, some methods~\cite{DBLP:conf/cvpr/CubukZMVL19, DBLP:conf/icml/HoLCSA19, Lin_2019_ICCV, DBLP:conf/nips/LimKKKK19} attempted to automate learning data augmentation policies. AutoAugment~\cite{DBLP:conf/cvpr/CubukZMVL19} models the policy search problem as a sequence prediction problem, and uses an RNN  controller to predict the policy. Reinforcement learning (RL) is exploited to optimize the controller parameters. Though promising results are achieved, AutoAugment is extremely costly (e.g., 15000 GPU hours on ImageNet) due to the low efficiency of RL and multi-pass training. PBA~\cite{DBLP:conf/icml/HoLCSA19} proposes to use population based training to achieve greater efficiency than AutoAugment, and evaluates on  small datasets like CIFAR-10 and SVHN. OHL-Auto-Aug~\cite{Lin_2019_ICCV} employs online hyper-parameter learning in searching for an auto-augmentation strategy, leading to a speedup over AutoAugment of 60$\times$  on CIFAR-10 and 24$\times$  on ImageNet. Fast AutoAugment (Fast AA)~\cite{DBLP:conf/nips/LimKKKK19} treats policy search as a density matching problem and applies Bayesian optimization to learn the policy, and achieves, e.g., $33\times$ speedup over AutoAugment on ImageNet. Although Fast AA achieves encouraging results, its cost is still high on large datasets. E.g., 450 GPU (Tesla V100) hours on ImageNet. 

Since our  DADA is inspired by the differentiable neural architecture search ~\cite{DBLP:conf/iclr/LiuSY19, DBLP:conf/iclr/XieZLL19, Dong_2019_CVPR}, we briefly review these methods here. DARTS~\cite{DBLP:conf/iclr/LiuSY19} first constructs a super-network of all possible operations and controls the super-network with architecture parameters. It then models neural architecture search as a bi-level optimization problem and optimizes  architecture parameters through stochastic gradient descent. In order to remove the bias of DARTS for operation selection, SNAS~\cite{DBLP:conf/iclr/XieZLL19} and GDAS~\cite{Dong_2019_CVPR} add stochastic factors to the super-network and utilize the Gumbel-Softmax gradient estimator~\cite{DBLP:conf/iclr/MaddisonMT17, DBLP:conf/iclr/JangGP17} to estimate the gradient. Our baseline method, which also uses the Gumbel-Softmax gradient estimator to optimize the data augmentation policy, is most motivated by these methods.

\section{Differentiable Automatic Data Augmentation (DADA)}
\label{sec:method}
We first introduce the search space of DADA in Section~\ref{subsec:search_space}. Then we model DA optimization as Monte Carlo sampling of DA policy in Section~\ref{subsec:sampling_distrition}. After that, we introduce the Gumbel-Softmax~\cite{DBLP:conf/iclr/JangGP17, DBLP:conf/iclr/MaddisonMT17} as a relaxation of the categorical distribution in Section~\ref{subsec:gumbel}. In Section~\ref{subsec:relax}, we introduce an unbiased gradient estimator with small variance, RELAX~\cite{DBLP:conf/iclr/GrathwohlCWRD18}, to compute gradients in our DADA framework.
Finally, we propose an efficient \emph{one-pass} optimization strategy to jointly optimize the network weights and the DA parameters in Section~\ref{subsec:bilevel}.

\subsection{Search Space}
\label{subsec:search_space}

	Following 
	Fast AutoAugment~\cite{DBLP:conf/nips/LimKKKK19},  a DA policy $P$ contains several sub-policies $s_i \ (1 \leq i \leq |P|)$. Each sub-policy $s$ includes 
	$k$ image operations (
	e.g. flipping, rotation)
	$\{\bar{O}_i^s(x;p_i^s, m_i^s)|1 \leq i \leq k\}$ 
	which are 
	applied in sequence. 
	Each 
	operation $\bar{O}_i^s$ is applied to the image $x$ 
	with 
	two 
	continuous parameters: $p_i^s$ (the probability of applying the operation) and $m_i^s$ (the magnitude of the operation): 
	\begin{gather}
	\bar{O}_i^s(x;p_i^s, m_i^s) = 
	\begin{cases}
	O_i^s(x;m_i^s) &\text{: with probability } p_i^s, \\
	x &\text{: with probability } 1-p_i^s. \\
	\end{cases}
	\end{gather}
	Therefore the sub-policy $s$ can be represented by a composition of operations:
	\begin{align}
	s(x;\mathbf{p}^s, \mathbf{m}^s) = & \bar{O}^s_k(x;p^s_k, m^s_k)\circ\bar{O}^s_{k-1}(x;p^s_{k-1}, m^s_{k-1})\circ\cdots\circ\bar{O}^s_{1}(x;p^s_1, m^s_1) \notag \\
	= & \bar{O}^s_k(\bar{O}^s_{k-1}(...\bar{O}^s_{1}(x;p_1^s,m_1^s);p_{k-1}^s, m_{k-1}^s);p_k^s,m_k^s). \label{eq:subpolicy}
	\end{align}

\subsection{Policy Sampling from a Joint Distribution}
\begin{figure}[t]
	\centering
	\includegraphics[width=0.95\linewidth]{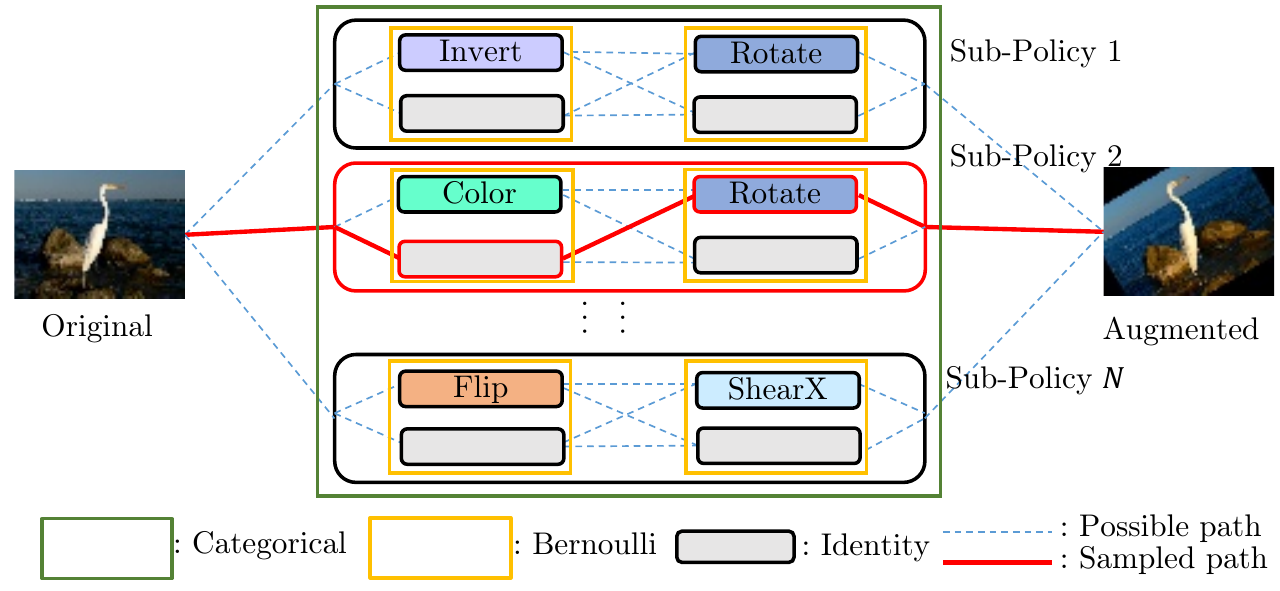}
	\label{subfig:methods}
	\caption{The framework of DADA. The sub-policies and operations are sampled from Categorical and Bernoulli distributions respectively.}
\end{figure}
\label{subsec:sampling_distrition}
We model the sub-policy selection and operation applying as sampling from Categorical and Bernoulli distributions, respectively. Then the DA optimization is modelled as a Monte Carlo gradient estimate problem from two distributions.
\subsubsection{Sub-Policy Sampling}
Let $\mathcal{O}$ be the set of all image pre-processing operations in our search space. Then the candidate $N$ sub-policies $\mathcal{S}$ are all the combinations of $k$  operations in  $\mathcal{O}$. 
To
determine which sub-policies should be selected, we sample the sub-policy from a Categorical distribution $p(c|\pi)$  (where we can sample a one-hot random variable) with probabilities $\pi$. Probabilities are computed as a softmax over parameters $\alpha=\alpha_{1:N}$ defining preference for sub-policies:
\begin{gather} 
\bar{s}(x) =\sum_{s\in\mathcal{S}}c_ss(x),
\quad 
c \sim p(c|\pi), \ \pi_{s} = \frac{exp(\alpha_{s})}{\sum_{s'\in\mathcal{S}}exp(\alpha_{s'})},
\quad
\text{for} \  s = 1,\dots ,N. \label{eq:categorical} 
\end{gather}
After $\alpha$ optimized, we can select the sub-policies with the top probabilities as the final DA policy $P$. 
Therefore, to optimize the DA policy, our task becomes optimizing the parameters $\alpha$ of the sub-policy sampling distribution.

\subsubsection{Operation Sampling}
Given a particular sub-policy choice, the constituent operations are executed or not according to a Bernoulli distribution sample. 
We start from the simple example: a sub-policy $s$ contains only one operation $O_1^s$. Then,
the image operation $O$ (we omit the superscript and subscript of $O_1^s$ for simplicity) with application probability $\beta$ and magnitude $m$ can be represented:
\begin{gather}
s(x) = \bar{O}(x;m) = b O(x; m) + (1-b) x, 
\quad b \sim \text{Bernoulli}(\beta).  \label{eq:bernoulli}
\end{gather}
To generalize Eq. (\ref{eq:bernoulli}) to  a sub-policy with multiple operations,
{we formulate the composition of $O_i$ and $O_{i-1}$ in Eq.~(\ref{eq:subpolicy}) as below:}
\begin{align}
\bar{O}_{i}(x;\beta_{i},m_{i}) \circ \bar{O}_{i-1}(x;\beta_{i-1},m_{i-1}) = 
& ~b_ib_{i-1} O_{i}(O_{i-1}(x;m_{i-1});m_{i}) \notag \\
& + b_i(1-b_{i-1}) O_{i}(x;m_i) \notag \\
& + (1-b_{i})b_{i-1} O_{i-1}(x;m_{i-1}) \notag \\
& + (1-b_{i})(1-b_{i-1})x
.  
\end{align}
where $ b_{i-1} \sim \text{Bernoulli}(\beta_{i-1})$,
$b_i \sim \text{Bernoulli}(\beta_i)$.

\subsubsection{Optimization Objective}
With the above formulations, the DA policy can be sampled from a joint distribution $p(c, b|\alpha, \beta) = p(b|\beta,c)p(c|\alpha)$ (we use $p(c|\alpha)$ to describe Eq.~(\ref{eq:categorical})). Therefore, our objective can be represented as:
\begin{align}
\mathbb{E}_{c, b \sim p(c, b|\alpha, \beta)}[\text{Reward}(c, b)] = \mathbb{E}_{c, b \sim p(c, b|\alpha, \beta)}[\mathcal{L}_{w}(c, b)]. \label{eq:opti_obj}
\end{align}
where $\mathcal{L}_{w}(c, b)$ is the \emph{validation-set loss} achieved by the network (with weights $w$) which is trained using the DA policy \{$c, b$\} on training set. Unlike AutoAugment which uses validation \emph{accuracy} as reward, we use the validation \emph{loss} to facilitate gradient computation.

\subsection{Differentiable Relaxation with Gumbel-Softmax}
\label{subsec:gumbel}
{To learn the data augmentation policy, we need to estimate the gradient of the objective w.r.t. the parameters  \{$\alpha, \beta$\} of the categorical (sub-policy) and bernoulli (operation) distributions.
In this section, we use Gumbel-Softmax reparameterization trick~\cite{DBLP:conf/iclr/JangGP17} (a.k.a. concrete distribution~\cite{DBLP:conf/iclr/MaddisonMT17}) to reparameterize the parameters \{$\alpha, \beta$\} to be differentiable. Then we detail the estimation of
the gradient of magnitudes $m$ using the straight-through gradient estimator~\cite{DBLP:journals/corr/BengioLC13}.}

\subsubsection{Differentiable Relaxation of Sub-Policy Selection}
The Categorical distribution is not differentiable w.r.t. the parameter $\alpha$, therefore we use the Gumbel-Softmax reparameterization trick to achieve the differentiable relaxation. This reparameterization is also used in network architecture search, e.g  SNAS~\cite{DBLP:conf/iclr/XieZLL19} and GDAS~\cite{Dong_2019_CVPR}. 
With the Gumbel-Softmax reparameterization, Eq. (\ref{eq:categorical}) becomes:
\begin{gather}
\bar{s}(x) = \sum_{s\in\mathcal{S}}c_{s}s(x), \notag \\
\quad
c \sim \text{RelaxCategorical}(\alpha, \tau) = \frac{\exp((\alpha_{s}+g_{s})/\tau)}{\sum_{s'\in\mathcal{S}}\exp((\alpha_{s'}+g_{s'})/\tau)},
\label{eq:backward}
\end{gather}
where $g = -\log(-\log(u))$ with $u \sim \text{Uniform}(0, 1)$, and $\tau$ is the temperature of softmax function.
In our implementation, the straight-through gradient estimator is applied: the backward pass uses differentiable variables as Eq.~(\ref{eq:backward}), the forward pass uses  discrete variables as shown:
\begin{gather}
\bar{s}(x) = \sum_{s\in\mathcal{S}} h_s s(x),
\quad
\text{where} \  h = \text{one\_hot}(\text{argmax}_{s}(\alpha_{s}+g_{s})). \label{eq:forward}
\end{gather}

\subsubsection{Differentiable Relaxation of Augmentation Operator Sampling}
Similar to the Categorical distribution, the Bernoulli distribution is not differentiable w.r.t. $\beta$. We apply the same reparameterization trick to the Bernoulli distribution: 
\begin{gather}
\text{RelaxBernoulli}(\lambda, \beta) = \sigma((\log{\frac{\beta}{1-\beta}}+\log{\frac{u}{1-u}})/\lambda), \ u \sim \text{Uniform}(0, 1). \label{eq:relaxbernoulli}
\end{gather}



Similar to 
Eq.~(\ref{eq:forward}), we only execute the 
operation if $b>0.5$ but backpropagate using the gradient estimated by 
Eq.~(\ref{eq:relaxbernoulli}).

\subsubsection{Optimization of Augmentation Magnitudes}
Different from optimizing the discrete distribution parameters, we optimize augmentation magnitudes by approximating their gradient.
Since some operations (e.g. flipping and rotation) in our search space 
are not differentiable w.r.t. the magnitude, 
we apply the straight-through gradient estimator~\cite{DBLP:journals/corr/BengioLC13} to optimize the magnitude.
{For an image $\hat{x}=s(x)$ augmented by sub-policy $s$,} we approximate the gradient of magnitude (of the sampled operations) w.r.t. each pixel of the augmented image: 
\begin{align}
\frac{\partial\hat{x}_{i,j}}{\partial m} = 1. \label{eq:magnitude}
\end{align}
Then the gradient of the magnitude w.r.t. our objective $L$ can be calculated as:
\begin{align}
    \frac{\partial L}{\partial m} = \sum_{i,j} \frac{\partial L}{\partial \hat{x}_{i,j}} \frac{\partial \hat{x}_{i,j}}{\partial m} = \sum_{i,j} \frac{\partial L}{\partial \hat{x}_{i,j}}.
\end{align}



\subsection{RELAX Gradient Estimator}
\label{subsec:relax}
Although the above reparameterization trick make 
DA parameters
differentiable, its gradient estimate is biased \cite{DBLP:conf/iclr/JangGP17, DBLP:conf/iclr/MaddisonMT17}. To address this problem, we propose to use RELAX~\cite{DBLP:conf/iclr/GrathwohlCWRD18} estimator which provides unbiased gradient estimate. 
{We first describe the  RELAX estimator for  gradient estimation w.r.t distribution parameters.
Then, we introduce the use of RELAX estimator to relax the parameters of Categorical  $p(c|\alpha)$ and Bernoulli  $p(b|\beta)$ distributions to be differentiable.}

\subsubsection{Gradient Estimation}
Here, we consider how to estimate the gradient of parameters of distribution $p(q|\theta)$, which can represent either $p(c|\alpha)$ or $p(b|\beta)$ in our algorithm.
Unlike the Gumbel-Softmax estimator, the gradient of RELAX  cannot simply be achieved 
by the backward of loss function. 
Furthermore, the RELAX estimator even does not require 
the loss function $\mathcal{L}$ to be differentiated w.r.t.  the distribution parameters. 
It means we do not need to
apply continuous relaxation the forward stage like Eq.~(\ref{eq:categorical}) or Eq.~(\ref{eq:bernoulli}).
Instead,
the RELAX estimator requires 
a differentiable neural network $c_{\phi}$ which is a surrogate of loss function, where $\phi$ are the parameters of neural network $c$.
To make $c_{\phi}$ be differentiable w.r.t. the distribution parameters, we need the same Gumbel-Softmax reparameterized distribution $p(z|\theta)$ for $p(q|\theta)$, 
where $\theta$ is the distribution parameter and $z$ is a sampled continuous variable.
$p(z|\theta)$ can be either the Gumbel-Softmax reparameterized distributions of $p(c|\alpha)$ or $p(b|\beta)$. 
Then, in the forward stage, the DA policy can also be sampled from $z \sim p(z|\theta)$ but only forward the policy using deterministic mapping $q=H(z)=b \sim p(b|\theta)$ to guarantee the probability distribution curve is the  same.
Furthermore, since RELAX applies the
control variate at the relaxed input $z$, we also need a relaxed input conditioned on the discrete variable $q$, denoted as $\tilde{z} \sim p(z|q, \theta)$.
Then the gradient estimated by RELAX can be expressed as:
\begin{gather}
\nabla_{\theta} \mathcal{L} = [\mathcal{L}(q) - c_{\phi}(\tilde{z})] \nabla_{\theta} \log{p(q|\theta)} + 
\nabla_{\theta}c_{\phi}(z) -
\nabla_{\theta}c_{\phi}(\tilde{z}), \\
q=H(z), z \sim p(z|\theta), \tilde{z} \sim p(z|q, \theta). \notag
\end{gather}
To achieve a small variance estimator,
the gradient of parameters $\phi$ can be computed as Eq.~(\ref{eq:phi}), which can be jointly optimized with the parameters $\theta$.
\begin{gather}
\nabla_{\phi}(\text{Variance}(\nabla_{\theta} \mathcal{L})) = \nabla_{\phi}(\nabla_{\theta} \mathcal{L})^2. \label{eq:phi}
\end{gather}

\subsubsection{Sub-Policy Sampling}
As shown in Eq. 
(\ref{eq:categorical}), the sub-policies can be sampled from the Categorical distribution. 
To 
apply the RELAX estimator, we first sample $z$ from the relaxed Categorical distribution $\text{RelaxCategorical}(\alpha, \tau)$ from 
Eq.~(\ref{eq:backward}), but we only forward the sub-policy with $c=\text{one\_hot}(\text{argmax}(z))$ 
in Eq.~(\ref{eq:forward}).
We further sample $\bar{z}$ conditioned on variable $c$ to control the variance:
\begin{gather}
\tilde{z}_{s} = \frac{\tilde{z}_{s}}{\sum_{s'\in\mathcal{S}}\tilde{z}_{s'}}, 
\notag \\
\text{where} \quad
\hat{z_i}  = \begin{cases}
-\log(-\log v_i )& c_i = 1 \\
-\log \left( -\frac{\log v_i}{p_i} - \log v_c \right) & c_i = 0 \\
\end{cases} ,
\ 
v \sim \text{Uniform}(0, 1).
\end{gather}

\subsubsection{Operation Sampling}
As shown in Eq. (\ref{eq:bernoulli}), the parameters $b$ can be sampled from Bernoulli distribution. To adapt to the RELAX gradient estimator, we also utilize the Gumbel-Softmax reparameterization trick for  $p(z|\beta)$ like Eq.~(\ref{eq:relaxbernoulli}), but only forward the operation with $b = \mathbb{I}(z>0.5)$, where $\mathbb{I}$ is the indicator function. 
To control the variance, we sample $\tilde{z}$ conditioned as the sampled parameters $b$:
\begin{gather}
\tilde{z} = \sigma((\log{\frac{\beta}{1-\beta}} + \log{\frac{v'}{1-v'}})/\tau), \notag
\\
\text{where} \quad
v' = \begin{cases}
v \cdot (1-p), b=0\\
v \cdot p + (1-p), b=1\\
\end{cases}
,	 
\ 
v \sim \text{Uniform}(0,1).
\end{gather}
As for the gradient of magnitudes, we approximate them following Eq.~(\ref{eq:magnitude}).
{Since we need to estimate the gradient of the parameters of joint distribution $p(c,b|\alpha, \beta)$, we  feed $c_{\phi}$ with the relaxed variables $c$ and $b$ in the same time for the surrogate of loss function.}

\subsection{Bi-Level Optimization}
\label{subsec:bilevel}
We have discussed the differentiable optimization  of the DA policy in Section 3.1-3.4. We now discuss the joint optimization of  DA policy and neural network.
{Clearly, it is very slow to sequentially iterate  1) the optimization of  DA policy,  2) neural network training  and  performance evaluation until converge. Ideally, we conduct 1) and 2) by training neural network once, i.e. \emph{one-pass} optimization. To achieve this, we propose the following  bi-level optimization strategy.}


Let $\mathcal{L}_{\text{train}}$ and $\mathcal{L}_{\text{val}}$ denote the training and  validation loss, respectively. 
The optimization objective is to find
the optimal parameters $d^{*}={\{\alpha^*, \beta^*, m^*, \phi^*\}}$ 
which
minimizes the validation loss $\mathcal{L}_{\text{val}}(w^{*})$. The weights $w^{*}$ are obtained by minimizing the expectation of 
training loss $w^{*} = \text{argmin}_{w}~ \mathbb{E}_{\bar{d} \sim p(\bar{d}|d)}[\mathcal{L}_{\text{train}}(w, \bar{d})]$
($m$ and $\phi$ are just the original parameters without sampling).
For simplicity, we drop
the sampling notation of the  training loss as  $\mathbb{E}[\mathcal{L}_{\text{train}}(w, {d})]$.
Therefore, our joint optimization objective can be represented as a bi-level problem:
\begin{gather}
\min~ \mathcal{L}_{\text{val}} (w^{*}(d)),\\
\text{s.t.} \quad w^{*}(d) = \text{argmin}_{w}~ \mathbb{E}[\mathcal{L}_{\text{train}}(w, d)].  \notag 
\end{gather}

Directly solving the above bi-level optimization problem would require repeatedly computing the model weights $w^*(d)$ as policy parameters $d$ are changed.
To avoid that,
we optimize $w$ and $d$ alternately through gradient descent. At step $k$, give the current data augmentation parameters $d_{k-1}$, we obtain $w_k$ by gradient descent w.r.t. expectation of training loss $\mathbb{E}[\mathcal{L}_{\text{train}}(w_{k-1},d_{k-1})]$. Then, we approximate the above bi-level objective through a single virtual gradient step:
\begin{gather}
\mathcal{L}_{\text{val}}(w_k - \zeta \nabla_{w}\mathbb{E}[\mathcal{L}_{\text{train}}(w_k, d_{k-1})]). \label{eq:val}
\end{gather}
Then, the gradient of Eq. (\ref{eq:val}) w.r.t. $d$ is (with the step index $k$ removed for simplicity):
\begin{gather}
-\zeta\nabla_{d, w}^{2}\mathbb{E}[\mathcal{L}_{\text{train}}(w, d)]\nabla_{w'}\mathcal{L}_{\text{val}}(w'),
\end{gather}
where $w' = w - \zeta \nabla_{w}\mathbb{E}[\mathcal{L}_{\text{train}}(w, d)]$.
The gradient is expensive to compute, therefore we use the  finite difference approximation. Let $\epsilon$ be a small scalar, $w^{+} = w+\epsilon\nabla_{w'}\mathcal{L}_{\text{val}}(w')$ and $w^{-} = w-\epsilon\nabla_{w'}\mathcal{L}_{\text{val}}(w')$. Then the gradient can be computed as:
\begin{gather}
-\zeta \frac{\nabla_{d}\mathbb{E}[\mathcal{L}_{\text{train}}(w^{+}, d)] - \nabla_{d}\mathbb{E}[\mathcal{L}_{\text{train}}(w^{-}, d)]}{2\epsilon}.
\end{gather}
As for the gradients $\nabla_{d}\mathbb{E}[\mathcal{L}_{\text{train}}(w^{+}, d)]$ and $\nabla_{d}\mathbb{E}[\mathcal{L}_{\text{train}}(w^{-}, d)]$, they can be estimated by the techniques mentioned in Section \ref{subsec:gumbel} and Section \ref{subsec:relax}. Specifically, we compute those two gradient with the same sampling sub-policy  to make the difference of the two gradients more reliable.
For the hyper-parameters $\zeta$ and $\epsilon$, we follow the settings of another bi-level optimization DARTS \cite{DBLP:conf/iclr/LiuSY19}, where $\zeta = \{\text{learning rate of} \ w\}$ and $\epsilon = 0.01/\|\nabla_{w'}\mathcal{L}_{\text{val}}(w')\|_2$.

\section{Experiments}
\label{sec:experiments}
We compare our method with the effective baseline data augmentation method,  Cutout~\cite{DBLP:journals/corr/abs-1708-04552}, and the augmentation policy learners:  AutoAugment~\cite{DBLP:conf/cvpr/CubukZMVL19} (AA), Population Based Augmentation~\cite{DBLP:conf/icml/HoLCSA19} (PBA), Fast AutoAugment~\cite{DBLP:conf/nips/LimKKKK19} (Fast AA), and OHL Auto-Aug~\cite{Lin_2019_ICCV} (OHL AA) on the CIFAR-10~\cite{krizhevsky2009learning}, CIFAR-100~\cite{krizhevsky2009learning}, SVHN~\cite{netzer2011reading} and ImageNet~\cite{ILSVRC15} datasets.

\subsection{Settings}
\subsubsection{Search Space}
For the search space, we follow AA for a fair comparison.
Specifically, 
our search space contains the same $15$ 
data augmentation operations
as PBA \cite{DBLP:conf/icml/HoLCSA19}, which is also the same as AA and Fast AA except that the SamplePairing~\cite{DBLP:journals/corr/abs-1801-02929} is removed.
Following AA, our sub-policy consists of two DA operations ($k=2$), and the policy  consists of $25$ sub-polices.

\subsubsection{Policy Search}
Following \cite{DBLP:conf/cvpr/CubukZMVL19, DBLP:conf/icml/HoLCSA19, DBLP:conf/nips/LimKKKK19},
we search the DA policies on the reduced datasets and evaluate on the full datasets. 
Furthermore, we split half of the reduced datasets as training set, and the remaining half as validation set for the data augmentation search.
In the search stage, we search the policy parameters for $20$ epochs on the reduced datasets.
We use the Adam optimizer 
for the policy parameters {$d=\{\alpha, \beta, m, \phi\}$} optimization with learning rate $\eta_{d} = 5 \times 10^{-3}$, momentum $\beta = (0.5, 0.999)$ and weight decay $0$.
For the optimization of neural network parameters, 
we use momentum SGD as optimizer, with the same hyper-parameters  as  evaluation stage {except the batch size and the initial learning rate.} 
In the search stage, we only apply the {data augmentation policy to} 
training examples, and  set the batch size to $128$ for 
CIFAR-10  and $32$ for other  datasets.
The initial learning rate is set according to the batch size by the linear rule.
We set $\tau$ to $0.5$ and use a two layer fully connected neural network with $100$ hidden units for $c_{\phi}$.
Parameters $\alpha$ are initialized to $10^{-3}$. Parameters $\beta$ and magnitudes $m$ are all initialized to $0.5$.
\subsubsection{{Policy Evaluation}}
We use the official 
publicly available
code of Fast AA to {evaluate the searched DA policies} for a fair comparison. 
Following Fast  AA \cite{DBLP:conf/nips/LimKKKK19} and AA \cite{DBLP:conf/cvpr/CubukZMVL19}, we use the same hyper-parameters for policy evaluation: 
	weight decay, learning rate, batch size, training epoch. 

\begin{table}[t]
	\centering
	\caption{
		GPU hours spent on DA policy search and corresponding test error (\%). We use Wide-ResNet-28-10 model for CIFAR-10, CIFAR-100 and SVHN, and ResNet-50 for ImageNet. We use the \textit{Titan XP} to estimate the search cost as PBA. AA and Fast AA reported the search cost on \textit{Tesla P100} and \textit{Tesla V100} GPU  respectively. $^*:$ estimated.
		}

	\begin{subtable}[h]{0.49\linewidth}
		\caption{GPU hours}
		\resizebox{\linewidth}{!}{
		\begin{tabular}{lccccc}
			\toprule
			Dataset & AA~\cite{DBLP:conf/cvpr/CubukZMVL19}    & PBA~\cite{DBLP:conf/icml/HoLCSA19}   & Fast AA~\cite{DBLP:conf/nips/LimKKKK19} & OHL AA~\cite{Lin_2019_ICCV} & \textbf{DADA} \\
			\midrule
			CIFAR-10 & 5000  & 5     & 3.5   & 83.4$^*$ & 0.1 \\
			CIFAR-100 & -     & -     & -     & - & 0.2 \\
			SVHN  & 1000  & 1     & 1.5   & - & 0.1 \\
			ImageNet & 15000 & -     & 450  & 625$^*$ & 1.3 \\
			\bottomrule
		\end{tabular}%
		}
	\end{subtable}
	\begin{subtable}[h]{0.49\linewidth}
		\caption{Test set error rate (\%)}
		\resizebox{\linewidth}{!}{
		\begin{tabular}{cccccc}
			\toprule
			Dataset & AA~\cite{DBLP:conf/cvpr/CubukZMVL19}    & PBA~\cite{DBLP:conf/icml/HoLCSA19}   & Fast AA~\cite{DBLP:conf/nips/LimKKKK19} & OHL AA~\cite{Lin_2019_ICCV} & \textbf{DADA} \\
			\midrule
			CIFAR-10 & 2.6   & 2.6   & 2.7  & 2.6 & 2.7 \\
			CIFAR-100  & 17.1  & 16.7  & 17.3 & - & 17.5 \\
			SVHN  & 1.1   & 1.2   & 1.1  & - & 1.2 \\
			ImageNet & 22.4  & -     & 22.4 & 21.1 & 22.5 \\
			\bottomrule
		\end{tabular}%
		}
	\end{subtable}
	\label{tab:searchtime}%
\end{table}%

\subsection{Results}
\begin{table}[t]
	\centering
	\caption{{CIFAR-10 and CIFAR-100 test error rates (\%).}}
	\resizebox{\linewidth}{!}{			
		\begin{tabular}{ll|ccccc|c}
			\toprule
			Dataset & Model & Baseline & Cutout~\cite{DBLP:journals/corr/abs-1708-04552} & AA~\cite{DBLP:conf/cvpr/CubukZMVL19}    & PBA~\cite{DBLP:conf/icml/HoLCSA19}   & Fast AA~\cite{DBLP:conf/nips/LimKKKK19} & \textbf{DADA} \\
			\midrule
			\midrule
			CIFAR-10 & Wide-ResNet-40-2 & 5.3   & 4.1   & 3.7   & -     & 3.6   & 3.6 \\
			CIFAR-10 & Wide-ResNet-28-10 & 3.9   & 3.1   & 2.6   & 2.6   & 2.7   & 2.7 \\
			CIFAR-10 & Shake-Shake(26 2x32d) & 3.6   & 3.0     & 2.5   & 2.5   & 2.7   & 2.7 \\
			CIFAR-10 & Shake-Shake(26 2x96d) & 2.9   & 2.6   & 2.0     & 2.0     & 2.0     & 2.0 \\
			CIFAR-10 & Shake-Shake(26 2x112d) & 2.8   & 2.6   & 1.9   & 2.0     & 2.0     & 2.0 \\
			CIFAR-10 & PyramidNet+ShakeDrop & 2.7   & 2.3   & 1.5   & 1.5   & 1.8   & 1.7 \\
			\midrule
			CIFAR-100 & Wide-ResNet-40-2 & 26.0    & 25.2  & 20.7  & -     & 20.7  & 20.9 \\
			CIFAR-100 & Wide-ResNet-28-10 & 18.8  & 18.4  & 17.1  & 16.7  & 17.3  & 17.5 \\
			CIFAR-100 & Shake-Shake(26 2x96d) & 17.1  & 16.0    & 14.3  & 15.3  & 14.9  & 15.3 \\
			CIFAR-100 & PyramidNet+ShakeDrop & 14.0    & 12.2  & 10.7  & 10.9  & 11.9  & 11.2 \\
			\bottomrule
		\end{tabular}%
	}
	\label{tab:cifar}%
\end{table}%

\subsubsection{CIFAR-10 and CIFAR-100}
Both CIFAR-10 and CIFAR-100 have $50,000$ training examples. { Following \cite{DBLP:conf/cvpr/CubukZMVL19,DBLP:conf/icml/HoLCSA19,DBLP:conf/nips/LimKKKK19}, we conduct DA optimization on the reduced CIFAR-10 dataset ($4,000$ randomly selected examples) and evaluate the trained policies on the full CIFAR-10 test set. \cite{DBLP:conf/cvpr/CubukZMVL19,DBLP:conf/icml/HoLCSA19, DBLP:conf/nips/LimKKKK19} use the discovered policies from CIFAR-10 and evaluate these policies on CIFAR-100. Since our DADA is much more efficient than other methods, thus, we conduct both the search (using a reduced dataset of $4,000$ randomly selected examples) and evaluation on CIFAR-100.} 
Following \cite{DBLP:conf/cvpr/CubukZMVL19,DBLP:conf/icml/HoLCSA19, DBLP:conf/nips/LimKKKK19}, we
search the DA policy 
using a Wide-ResNet-40-2 network ~\cite{DBLP:conf/bmvc/ZagoruykoK16}
and evaluate the 
searched
policy using 
Wide-ResNet-40-2~\cite{DBLP:conf/bmvc/ZagoruykoK16}, Wide-ResNet-28-10~\cite{DBLP:conf/bmvc/ZagoruykoK16}, Shake-Shake~\cite{DBLP:conf/iclr/Gastaldi17} and PyramidNet+ShakeDrop~\cite{DBLP:journals/access/YamadaIAK19}. 

From the results in Table~\ref{tab:searchtime} and \ref{tab:cifar}, we can see that DADA requires significantly less computation than the competitors, while providing comparable error rate. For example, we require only $0.1$ GPU hours for  policy search on  reduced CIFAR-10, which is at least one order of magnitude  faster than AA ($50,000\times$) and Fast AA ($35\times$). Similar to CIFAR-10, we achieve very competitive performance  on  CIFAR-100 yet with much less searching cost.  
Despite the lower error rates of OHL AA, 
 OHL AA is not directly comparable to other methods since it uses a larger and dynamic search space.

\subsubsection{SVHN}
To verify the generalization ability of our search algorithm for different datasets, we further conduct experiments with a larger dataset: SVHN. 
 SVHN dataset has $73,257$ training examples (`core training set'), $531,131$ additional training examples, and $26,032$ testing examples.
Following
AA and Fast AA, we also search the DA  policy with the reduced SVHN dataset, which has $1,000$ 
randomly selected training samples from the core training set.
{Following AA, PBA and Fast AA}, we evaluate the 
learned DA policy performance with the full SVHN training data.
Unlike AA, we use the Wide-ResNet-28-10 \cite{DBLP:conf/bmvc/ZagoruykoK16} architecture 
in the search stage and evaluate the policy on the Wide-ResNet-28-10 \cite{DBLP:conf/bmvc/ZagoruykoK16} and Shake-Shake (26 2x96d) \cite{DBLP:conf/iclr/Gastaldi17}. 
Our results are shown in  Table~\ref{tab:svhn}.
As shown in Table~\ref{tab:svhn}, our DADA achieves
similar error rate to PBA,  slightly worse than AA and Fast AA. However, we only use 
	$0.1$ GPU hours 
	in the search stage.

\begin{table}[t]
	\centering
	\caption{{SVHN test error rates (\%).}}
	\begin{tabular}{l|ccccc|c}
		\toprule
		Model & Baseline & Cutout~\cite{DBLP:journals/corr/abs-1708-04552} & AA~\cite{DBLP:conf/cvpr/CubukZMVL19} & PBA~\cite{DBLP:conf/icml/HoLCSA19}   & Fast AA~\cite{DBLP:conf/nips/LimKKKK19} & \textbf{DADA} \\
		\midrule
		Wide-ResNet-28-10 & 1.5   & 1.3   & 1.1   & 1.2   & 1.1   & 1.2 \\
		Shake-Shake(26 2x96d) & 1.4   & 1.2   & 1.0  & 1.1  & -     & 1.1 \\
		\bottomrule
	\end{tabular}%
	\label{tab:svhn}%
\end{table}
\begin{table}[t]
	\centering
	\caption{{Validation set Top-1 / Top-5 error rate (\%) on ImageNet using ResNet-50 and DA policy search time (h). $^*$: Estimated. }}
	\begin{tabular}{c|cccc|c}
		\toprule
		 &Baseline &  AA~\cite{DBLP:conf/cvpr/CubukZMVL19}& Fast AA~\cite{DBLP:conf/nips/LimKKKK19}& OHL AA~\cite{Lin_2019_ICCV} & \textbf{DADA} \\
		\midrule
		Error Rate (\%) & 23.7 / 6.9 & ~22.4 / 6.2 & 22.4 / 6.3 & 21.1 / 5.7 & 22.5 / 6.5 \\
		Search Time (h)	 & 0 & 15000 & 450 & 625$^*$ & 1.3 \\
		\bottomrule
	\end{tabular}%
	\label{tab:imagenet}%
\end{table}%


\subsubsection{ImageNet}
Finally, we evaluate our algorithm on the ImageNet dataset.
We train the policy on the same 
ImageNet subset as
Fast AA, which consists of $6,000$ examples from a randomly selected $120$ classes.
We use
ResNet-50~\cite{DBLP:conf/cvpr/HeZRS16} for policy optimization
and report the 
performance trained with the full ImageNet dataset.
As shown in Table~\ref{tab:searchtime} and Table~\ref{tab:imagenet},
we achieve very competitive error rate against
AA and Fast AA while requiring only $1.3$ GPU hours in the search stage -- compared to $15000$ and $450$ hours respectively for these alternatives. 
Again, OHL AA ~\cite{Lin_2019_ICCV} uses a larger and dynamics search space, thus, it is not comparable to other methods. 


\subsection{DADA for Object Detection}
\begin{table}[t]
	\centering
	\caption{{Object detection bounding box (bb) and mask AP on COCO test-dev.}}
	\begin{tabular}{llccccccc}
		\toprule
		Method & Model & $\text{AP}^{\text{bb}}$    & $\text{AP}_{50}^{\text{bb}}$ & $\text{AP}_{75}^{\text{bb}}$ & $\text{AP}_{S}^{\text{bb}}$ & $\text{AP}_{M}^{\text{bb}}$ & $\text{AP}_{L}^{\text{bb}}$ & $\text{AP}^{\text{mask}}$\\
		\midrule
		\midrule
		RetinaNet & ResNet-50 (baseline) & 35.9  & 55.8  & 38.4  & 19.9  & 38.8  & 45.0 & - \\
		RetinaNet & ResNet-50 (DADA) & $\textbf{36.6}^{+0.7}$  & 56.8  & 39.2  & 20.2  & 39.7  & 46.0  & -\\
		\midrule
		Faster R-CNN & ResNet-50 (baseline) & 36.6  & 58.8  & 39.6  & 21.6  & 39.8  & 45.0  & -\\
		Faster R-CNN & ResNet-50 (DADA)  & $\textbf{37.2}^{+0.6}$  & 59.1  & 40.2  & 22.2  & 40.2  & 45.7 & -\\
		\midrule
		Mask R-CNN & ResNet-50 (baseline) & 37.4  & 59.3  & 40.7  & 22.0    & 40.6  & 46.3 & 34.2 \\
		Mask R-CNN & ResNet-50 (DADA) & $\textbf{37.8}^{+0.4}$ & 59.6  & 41.1  & 22.4  & 40.9  & 46.6 & $\textbf{34.5}^{+0.3}$\\
		\bottomrule
	\end{tabular}%
	\label{tab:coco}%
\end{table}%
The use of ImageNet pre-training  backbone networks is a common technique for object detection~\cite{DBLP:conf/iccv/LinGGHD17, DBLP:conf/nips/RenHGS15, DBLP:conf/iccv/HeGDG17}.
To improve the performance of object detection, people usually focus on designing a better detection pipeline, while paying less attention to improving the ImageNet pre-training backbones. It is interesting to investigate whether the backbones trained using our DADA can improve detection performance.
With our DADA algorithm, we have reduced 
the Top-1 error rate of ResNet-50 on ImageNet from $23.7\%$ to $22.5\%$. 
In this section, we further conduct experiments on object detection dataset MS COCO~\cite{DBLP:conf/eccv/LinMBHPRDZ14} with the better ResNet-50 model.
We adopt three mainstream detectors RetinaNet~\cite{DBLP:conf/iccv/LinGGHD17}, Faster R-CNN~\cite{DBLP:conf/nips/RenHGS15} and Mask R-CNN~\cite{DBLP:conf/iccv/HeGDG17} in our experiments.
For the same detector, we use the same setting as~\cite{DBLP:journals/corr/abs-1906-07155} except that the ResNet-50 model is trained with or without DADA policy.
From Table~\ref{tab:coco}, the performance of ResNet-50 trained with DADA policy is 
consistently
better than the ResNet-50 trained without DADA policy.
The results show that our learned DA policy also improves generalisation performance of downstream deep models that leverage the pre-trained feature.

\subsection{Further Analysis}
\subsubsection{Comparison with the Gumbel-Softmax Estimator}

\begin{figure}[t]
	
	\centering
	\begin{subfigure}{0.47\linewidth}
	\includegraphics[width=1.0\linewidth]{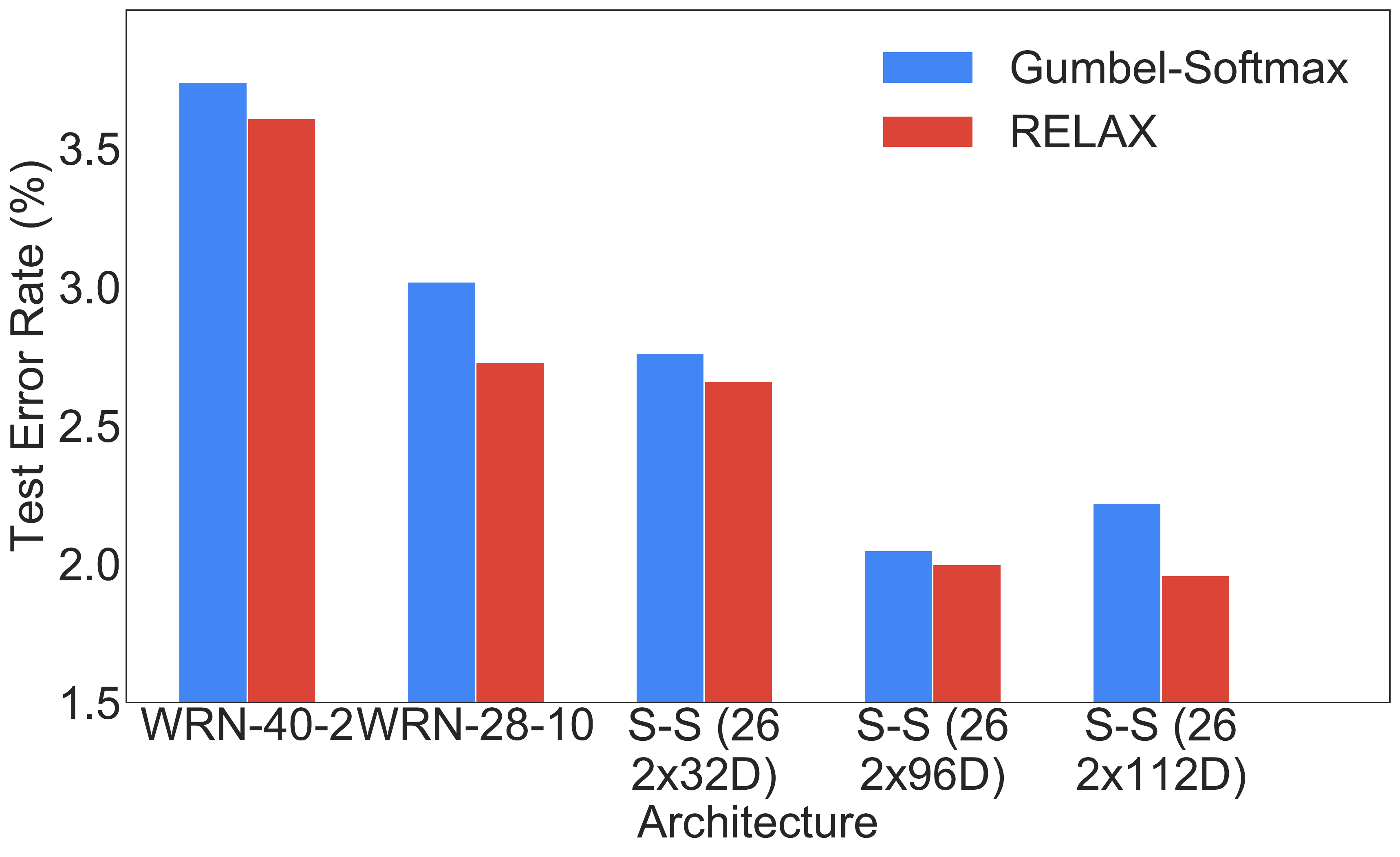}
		\caption{
			{Comparison with Gumbel-Softmax.}.
		}
		\label{subfig:relaxgumbelerror}
	\end{subfigure}
	\begin{subfigure}{0.51\linewidth}
		\includegraphics[width=1.0\linewidth]{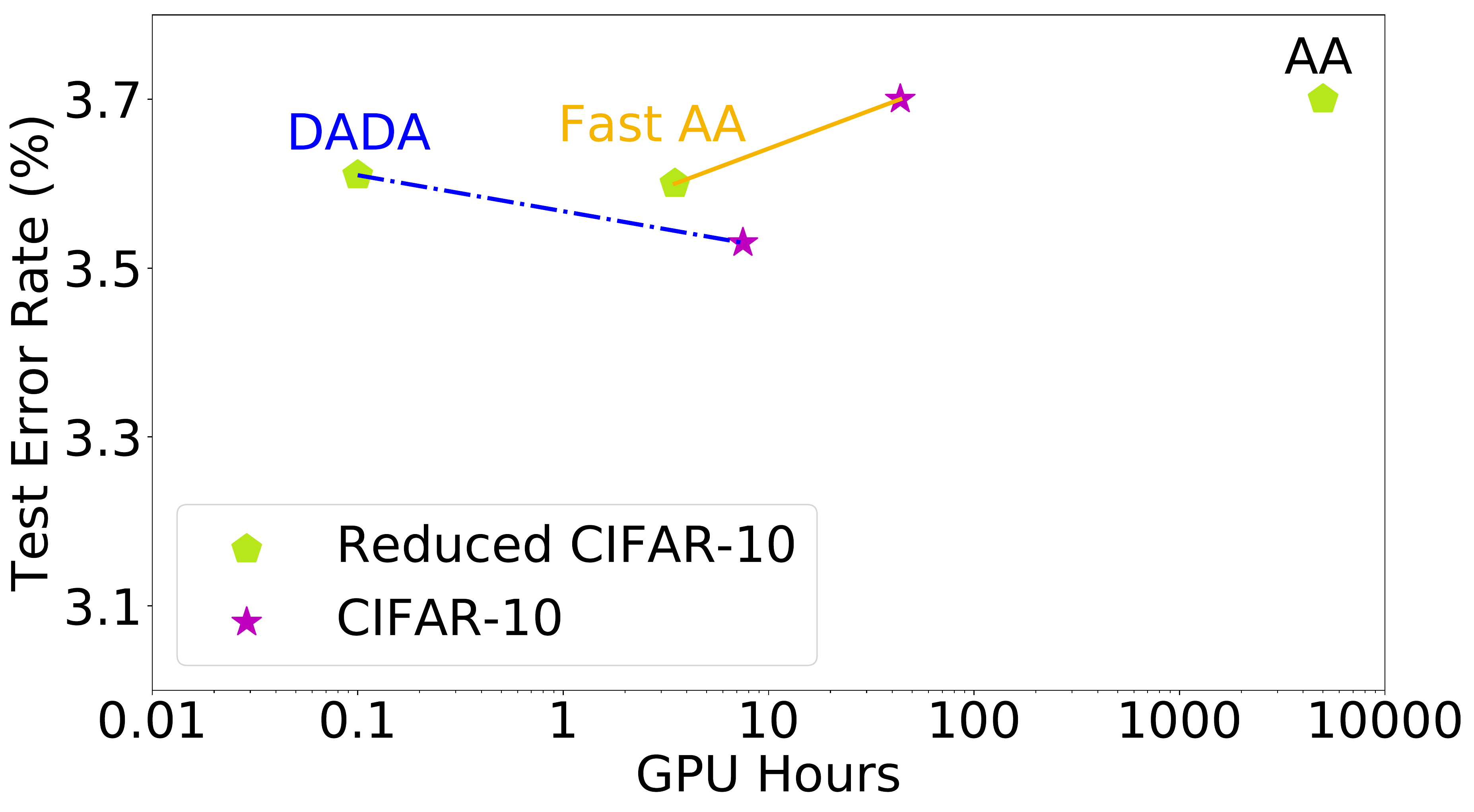}
		\caption{
			{Error rate vs policy search cost.}
		}
		\label{subfig:randomsearch}
	\end{subfigure}
	\caption{Additional Analysis on DADA.}
	\label{fig:ablationstudy}
\end{figure}

One technical contribution of this paper is the derivation of a RELAX estimator for DA policy search, which removes the bias of the conventional Gumbel-Softmax estimator. To evaluate the significance of this contribution, we conduct the search experiments for both estimators with the same hyper-parameters on the CIFAR-10 dataset. As we can see from Fig.~\ref{subfig:relaxgumbelerror}, the policy found using our RELAX estimator  achieves better performance on  CIFAR-10 compared with Gumbel-Softmax estimator.

\subsubsection{Search on the Full Dataset}
\begin{table}[t]
	\centering
	\caption{{The test error rate (\%) with DA policy learned on different training set.}}
	\begin{tabular}{llcc}
		\toprule
		Dataset & Model & Reduced CIFAR-10 & Full CIFAR-10 \\
		\midrule
		CIFAR-10 & Wide-ResNet-40-2 & 3.61  & 3.53 \\
		CIFAR-10 & Wide-ResNet-28-10 & 2.73  & 2.64 \\
		\bottomrule
	\end{tabular}%
	\label{tab:dataset}%
\end{table}%
We further 
evaluate the performance when we train the DA 
policy on the full dataset rather than on the reduced one, noting that this is feasible for the first time with DADA, due to its dramatically increased efficiency compared to alternatives. 
{We conduct DA policy search on both the reduced and full CIFAR-10 dataset with Wide-ResNet-40-2.}
As we can see from  Table~\ref{tab:dataset}, the policy searched on the full dataset works better than that on the reduced one evaluated on CIFAR-10.

We finally bring together some of these results and compare the speed accuracy trade-off provided by DADA, Fast AA and AA in Fig.~\ref{subfig:randomsearch} for CIFAR-10 on WRN-40-2 architecture.  Fast AA does not benefit from DA policy on the full CIFAR-10 dataset. However, DADA provides an excellent tradeoff, especially at low resource operating points.


\section{Conclusion}
\label{sec:conclusion}

In this work, we proposed Differentiable Automatic Data Augmentation (DADA)  for data augmentation policy learning. DADA relaxes the discrete policy selection process to be differentiable using Gumbel-Softmax. To achieve efficient and accurate optimization, we propose a \emph{one-pass} optimization strategy. In our differentiable optimization framework, we introduce an unbiased gradient estimator RELAX to achieve an accurate gradient estimation. Experimental results show that DADA achieves comparable image classification accuracy to state-of-the-art with at least one order of magnitude less search cost. 
DADA's greater efficiency makes it the first practical Auto-DA tool of choice that practitioners can use to optimize DA pipelines for diverse applications on desktop-grade GPUs. 


\section*{Acknowledgment}
This work is supported by National Natural Science Foundation of China under Grant 61673029. 

\clearpage
%
%

\bibliographystyle{splncs04}
\bibliography{augmentbib}
\end{document}